\documentclass[a4paper, 10 pt, conference]{ieeeconf}

\IEEEoverridecommandlockouts

\usepackage{graphics} 
\usepackage{graphicx}
\usepackage{epsfig} 
\usepackage{amsmath} 
\usepackage{amssymb}  
\usepackage{subfigure}
\usepackage{float}
\usepackage{url}
\usepackage{multirow}
\usepackage{subfigure}
\usepackage{wrapfig}
\usepackage{tabularx}

\usepackage{hyperref}
\newcommand\fnurl[2]{%
  \href{#2}{#1}\footnote{\url{#2}}%
}
\usepackage{color}
\usepackage{bm}
\usepackage{array}
\usepackage{booktabs}
\usepackage{algorithm,algorithmic}
\usepackage{siunitx}

\usepackage{gensymb}

\usepackage{lipsum}
\usepackage{makecell}
\usepackage{diagbox}
\usepackage{tikz}

\usepackage{colortbl}

\graphicspath{{./figs/}}

\newcolumntype{C}[1]{>{\centering}m{#1}}
\newcolumntype{L}{>{\centering\arraybackslash}m{0.7cm}}

\makeatletter
\newcommand{\thickhline}{%
    \noalign {\ifnum 0=`}\fi \hrule height 1pt
    \futurelet \reserved@a \@xhline
}
\newcolumntype{"}{@{\hskip\tabcolsep\vrule width 1pt\hskip\tabcolsep}}
\makeatother

\newcommand{\doctitle}{Don't Worry About the Weather:\\ Unsupervised Condition-Dependent Domain Adaptation}
\newcommand{\docsubtitle}{}
\title{\LARGE \bf\doctitle \small\break\docsubtitle}
\author{Horia Porav, Tom Bruls and Paul Newman
\thanks{Authors are from the Oxford Robotics Institute, University of Oxford, UK.
{\tt\small \{horia, tombruls, pnewman\}@robots.ox.ac.uk}}
}

\renewcommand{\baselinestretch}{0.95}

\begin{document}
\vspace{-30mm}
\maketitle
\vspace{-20mm}
\global\csname @topnum\endcsname 0
\global\csname @botnum\endcsname 0

\begin{abstract}
Modern models that perform system-critical tasks such as segmentation and localization exhibit good performance and robustness under ideal conditions (i.e. daytime, overcast) but performance degrades quickly and often catastrophically when input conditions change. In this work, we present a domain adaptation system that uses light-weight input adapters to pre-processes input images, irrespective of their appearance, in a way that makes them compatible with off-the-shelf computer vision tasks that are trained only on inputs with ideal conditions. No fine-tuning is performed on the off-the-shelf models, and the system is capable of incrementally training new input adapters in a self-supervised fashion, using the computer vision tasks as supervisors, when the input domain differs significantly from previously seen domains. We report large improvements in semantic segmentation and topological localization performance on two popular datasets, RobotCar and BDD.
\end{abstract}

\section{Introduction}\label{sec:introduction}

\begin{figure}
\centering
\noindent\includegraphics[width=\columnwidth]{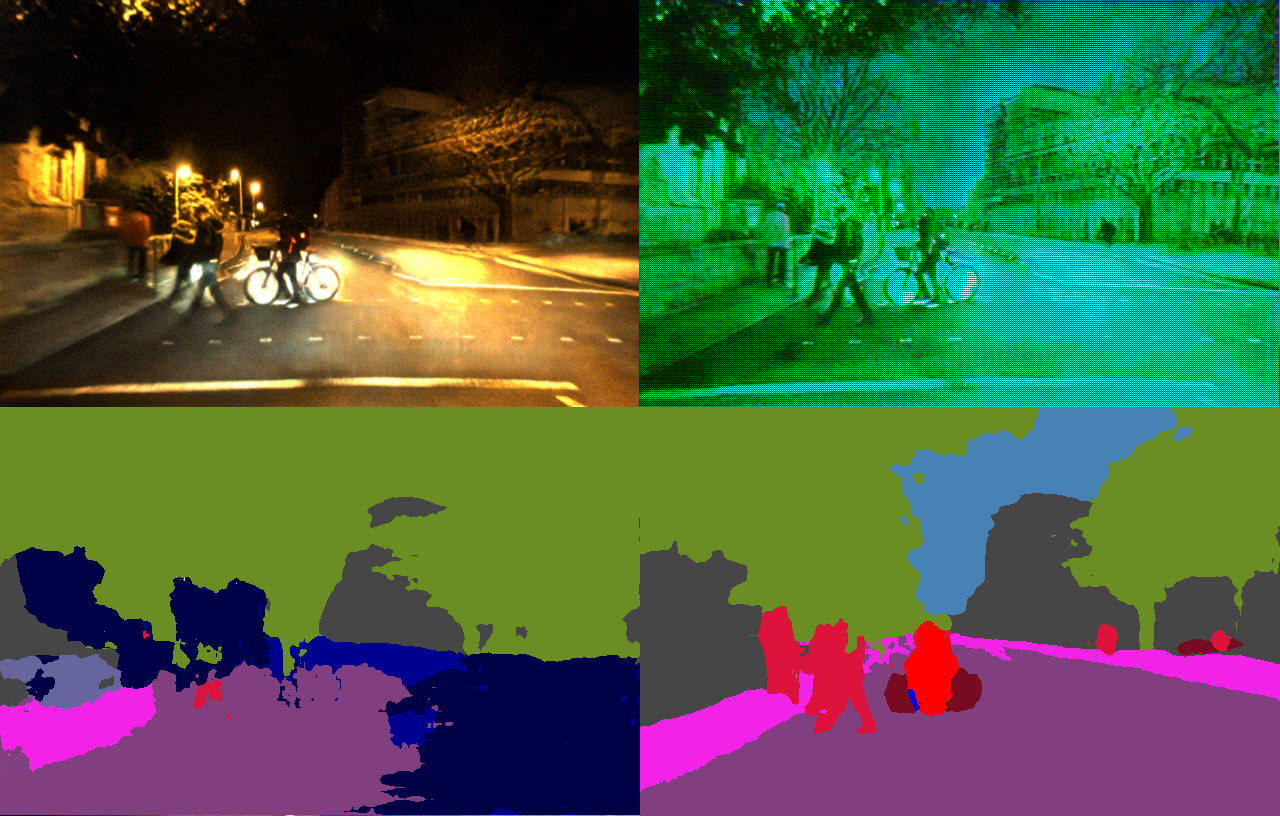} 
\caption{Our method allows off-the-shelf models to work with new, unseen domains, without any specific fine-tuning. In this example, we use our input adapter to allow a segmentation model trained using only daytime examples to work under night-time conditions. Top-left quadrant is the input night-time image, top-right quadrant is the output of our input adapter. The bottom-left quadrant is the output of the segmentation model applied on the original night-time input image. The bottom-right quadrant is the output of the segmentation model applied on the output of our adapter. The initial segmentation result is unusable, while the result obtained by running the model on the output of our domain adaptation pipeline accurately classifies roads, pavement, pedestrians, bicycles, vegetation and buildings.}
\label{fig:introfig}
\end{figure}

Robust Computer Vision is paramount to the prevalence of general-purpose robotics, and even more so in fields such as autonomous transportation, where failures may be catastrophic. Modern models that perform system-critical tasks such as segmentation, detection, localization and classification - either traditional heuristics-based or learned - exhibit good performance and robustness under ideal conditions (i.e. daytime, overcast) but performance degrades quickly and often catastrophically when input conditions change. To have their breakthrough, real-world systems must work under varying illumination, weather and noise conditions, and in the long term will need the ability to adapt to new, unseen domains without explicit supervision.

As a type of domain adaptation technique, domain unification is the holy grail of visual perception, theoretically allowing models trained on samples with limited heterogeneity to perform adequately on scenes that are well out of the distribution of the training data. Domain unification can be applied within the vast distribution of natural images \cite{vangooldarkmodel, Porav2018AdversarialTF, copycat}, between natural and synthetic images (computer-generated, whether through traditional 3D rendering or more modern GAN-based techniques) \cite{ma2018exemplar,cycada} and even between different sensor modalities \cite{crossmod1}. Additionally, domain unification can be implemented at different stages of a computer vision pipeline, ranging from direct approaches such as domain confusion \cite{deepdomainconfusion,WulfmeierIROS2017,crossdomainselfsup}, fine-tuning models on target domains \cite{vangooldarkmodel} or mixture-of-expert approaches \cite{mixofexp}, etc.

However, a major limiting factor for all these approaches is the scarcity of labelled multi-modal data, driven by the high cost of manual labelling. Most approaches attempt to solve this shortcoming by using 3D-rendered simulations that programatically provide ground-truth \cite{carla2017,synthia}, or by using unsupervised techniques that adapt models based on auxiliary or proxy tasks~\cite{deepdomainconfusion,WulfmeierIROS2017,crossdomainselfsup,learningfromsynthdata, vangooldarkmodel}. 

We propose a hybrid method, where multi-modal data is generated in an unsupervised fashion with approximated, high-quality ground truth, followed by supervised training of domain-adapters, for a battery of computer vision tasks, using this generated data and approximated ground truth. While this final step is supervised, the data used for supervision is itself created in an unsupervised fashion, making the entire pipeline unsupervised.
To do so, we start by generating multi-modal training data: from a database of image sequences categorized using the time of day and weather conditions at their moment of recording, we select a daytime, overcast, clear \textbf{reference} sequence. We leverage the fact that modern computer-vision pipelines perform excellent (e.g. $\geqslant 0.83$ mIOU on the Cityscapes multi-class segmentation validation set \cite{deeplabv3plus2018}) on inputs with ideal conditions, and thus we run this \textbf{reference} sequence through a set of off-the-shelf computer vision tasks and save the outputs as \emph{approximated} ground truth - these results are not 100\% identical to the real-world ground truth but they approximate it very well. 

Secondly, we apply style transfer to the \textbf{reference} sequence in order to produce a set of sequences which are structurally and geometrically identical to the \textbf{reference} sequence but differ in appearance. The applied style is sourced from many other sequences, each possessing a different appearance. This step is key to our approach - retaining the structure and geometry of the \textbf{reference} sequence while varying appearance means that the \emph{approximated} ground truth data is still valid: as an example, the same car but with varying appearances - once during daytime and once during nighttime - will still have the same ground truth footprint in a semantic segmentation map. At the end of these two steps we will have produced - in an unsupervised fashion - a set of sequences with varying appearances accompanied by task-specific ground truth data.

In step three, instead of training condition-dependent, separate task-specific models (e.g. a segmentation model for day-time, one for night-time etc.), we train lightweight condition-specific image adapters that are then used with vanilla, off-the-shelf task-specific models. The motivation for this is simple, but important: models that are invariant to input distributions are notoriously difficult to architect, parametrize and learn, while multi-model approaches such as mixture-of-experts do not scale well with the variance of the input distributions due to memory and runtime constraints. We tackle both these problems by training small, lightweight condition-specific convolutional input adapters, while a classifier-supervisor chooses the best adapter to be run, dependent on the distribution of the inputs. This approach adds minimal overhead to any off-the shelf task, benefits from parameter-counts that do not depend on the variances of the input distributions, and lends itself well to online learning of new conditions. Additionally, in contrast to the larger, task-specific models, the image adapter models tend to take up very little storage space and runtime memory and can be nearly-instantaneously loaded and re-loaded by the processing pipeline. 

The final stage of our approach allows a robot or vehicle to incrementally adapt to a new, unseen domain: if the condition of the input images does not match one that the system has been previously trained on, the unsupervised style transfer pipeline will select a model that is closest to the current condition, clone it, and fine-tune this cloned model to be able to change the style of the \textbf{reference} sequence so that it matches the style of the current input images. Afterwards, data generated using this new model will be used to train - in a supervised fashion -  an additional condition-specific image adapter that will allow upstream computer vision tasks to perform well on the new input image condition.

We benchmark our approach on two important tasks in computer vision and robotics: semantic segmentation and image retrieval/topological localization. This list is obviously not exhaustive, and the addition of extra supervisory tasks may lead to further improvements in performance.

Our main contributions include: 
\begin{itemize}
    \item Using cycle-consistency GANs to generate multi-condition training data with approximated ground truth for a battery of off-the-shelf computer vision tasks.
    \item Training input image adaptors by using the off-the-shelf computer vision models to generate a supervisory signal.
    \item Enabling online learning of new, unseen domains by leveraging the unsupervised data generation pipeline along with domains on which the data generation models have already been trained.
    \item Showing that training multiple lightweight adapter modules is better than training monolithic computer vision models that are invariant to input distributions.
\end{itemize} 

Our qualitative and quantitative results are presented in section \ref{sec:results}.

\section{Related Work}\label{sec:related-work}

\subsection{Computer Vision Tasks}
\subsubsection{\textbf{Semantic Segmentation}}\label{subsec:relsemseg}
Semantic segmentation is a key task in robotics, and modern approaches exhibit very good performance when input conditions are favourable. Deep convolutional models such as Deeplab V3+ \cite{deeplabv3plus2018}, SDN \cite{StackedDN} or PSPNet \cite{pspnet} achieve high class-mIOU figures ($\geqslant 80\%$) on benchmarks such as Cityscapes \cite{cordts2016cityscapes}, but their performance breaks down fast when their inputs change due to different weather conditions, seasons or times of day. For this work, we chose DeepLab V3+ as the reference model due to its excellent open-source implementation and availability of results on a number of popular benchmarks.
\subsubsection{\textbf{Topological Localization}}\label{subsec:reltopo}
Similarly, widely used topological localization frameworks such as FABMAP \cite{fabmap}, SeqSlam \cite{seqslam} or NetVLAD \cite{netvlad} achieve high recall and precision figures on clear, daytime images or when explicitly matching images with the same condition (e.g. winter-winter matching), but break down when the conditions of the locations to be matched differ. For this work, we chose NetVLAD as the reference topological localizer.

\subsection{Domain Adaptation}\label{subsec:reladapt}

\subsubsection{\textbf{Domain Confusion}}\label{subsec:reldomconf}

The most common approaches fall under the umbrella of domain confusion, making use of a discriminator that forces features extracted by an encoder to follow a similar distribution for both a source and a target domain \cite{deepdomainconfusion,WulfmeierIROS2017,crossdomainselfsup,learningfromsynthdata,Tsai2018LearningTA, semiweaklysupICCV2017}. The downside of these approaches is the lack of a direct loss for the target domain, which limits its upper bound on performance.

\subsubsection{\textbf{Style-Transfer}}\label{subsec:relstyletr}
Other approaches attempt to directly train computer vision models using synthetic data generated via style-transfer, or to directly adapt the input data to the target domain. Notable approaches include those of \cite{ma2018exemplar,cycada, Murez2018ImageTI, copycat} and  \cite{Porav2018AdversarialTF}. Generally, these methods seem to have the most promise of reducing the domain gap between real and synthetic images, hence our decision to generate training data using the approach of \cite{CycleGAN2017}.

In \cite{dundar2018domain}, a style-transfer pipeline is trained incrementally by feeding the segmentation map, obtained at a previous adaptation step, as an auxiliary input at each incremental step. The downside is that this type of  self-supervised approach assumes that high values in the softmax layer (using segmentation as an example) automatically correlate with higher prediction accuracies, whereas we only approximate ground truth labels once, on a reference, high-quality input sequence using models whose accuracies have actually been validated experimentally.

The closest to our approach is \cite{vangooldarkmodel}, where a semantic segmentation network is trained first with a day-time hand-labelled dataset, and then used to predict labels on intermediary datasets recorded at incremental types of twilight. The twilight images and estimated labels are then used to further fine-tune the segmentation model, which is finally used to segment night-time images. In contrast, our approach only computes approximated labels on the \textbf{reference} condition and uses the rest of the conditions as a guide for style transfer. However, the approach of \cite{vangooldarkmodel} could be used as a drop-in replacement for style transfer in the larger context of our framework. Similarly, \cite{vangool2semanticnight} trains segmentation models with a mixture of source domain images and synthetic images with the style of an incrementally-shifted target domain. The main difference between these approaches and our work is that instead of directly training or fine-tuning computer-vision tasks, we train lightweight input adapters while using the performance (loss) on these tasks as a supervisory signal.

Additionally, \cite{traincat} presents an approach where an encoder-decoder is trained to transform the appearance of input images to a reference appearance, but this is only benchmarked on scenes with small changes in appearance, in the context of 6-DOF localization.

\subsection{Online Learning}\label{subsec:relonline}
The authors of \cite{wulfmeier2017incremental} present an approach to incremental online domain adaptation, making use of unsupervised training by employing domain confusion at the level of encoder features from both target and source domains, while slowly shifting both domains through a range of incremental appearance changes (e.g. day to night). In our case, the unsupervised regimen is moved to the data-creation stage, with the generated data being used for supervised training of the input adapters, leading to better training stability and better performing models. Additionally, the authors present a method of reducing data storage requirements by approximating the feature distribution of the source data (or reference data, in our case) using a generative model, which could potentially be swapped for the reference sequence in our approach.  

Other approaches include map-management for lifelong learning \cite{mapmanagesiegwart} and adaptation on a domain manifold \cite{DLOW}. Our incremental learning pipeline follows the spirit of these works by always choosing to fine-tune a ‘seed’ model that was initially trained on data from a domain that is close to the current target domain.

\subsection{Expert Systems}\label{subsec:relexpert}
Finally, systems \cite{efficientcondlepetit,mixofexp} exist  that attempt to achieve invariance to the input conditions by running multiple sub-models in parallel and combining their outputs using a weighting or gating scheme to yield the desired result. One major issue with this type of approach is that runtime memory and processing power requirements increase linearly with the number of expert components used in parallel. While we also produce a number of input adapters that is proportional to the number of target domains, our classifier will choose only one input adapter to be run, per target domain, leading to a very small computational and memory footprint at runtime.


\section{Learning Condition-Dependent Representations}\label{sec:method}
\subsection{Synthetic Multi-Condition Data}\label{subsec:syntdata}

The first step in our approach is data generation. From the Oxford Robotcar Dataset \cite{RobotCarDatasetIJRR}, we select a \textbf{reference} sequence - one that is daytime, clear, overcast - and a number of traversals with difficult conditions - night, rain, snow etc. We use these conditions, along with a cycle-consistency architecture GAN \cite{CycleGAN2017}, to train generative models that can apply style transfer to the \textbf{reference} condition in order to create a number of synthetic sequences that maintain the structure and geometry of the \textbf{reference} condition - locations, shapes and topologies of both static and dynamic objects and of the overall scene - but exhibit variation in  appearance. In the following paragraphs we offer a succinct introduction to cycle-consistency GANs and how we use them to generate new data. The explanations offered here are in no way exhaustive, and interested readers should refer to the work of \cite{CycleGAN2017} for further details.

Following the work of \cite{CycleGAN2017}, we employ 2 generators: given an image $I_{\mathrm{A}}$ from domain $A$ (\textbf{reference}) and an image $I_{\mathrm{B}}$ from domain $B$ (night, rain, snow etc.), we use generator $G_{\mathrm{AB}}$ to translate an image style from domain A to domain B and generator $G_{\mathrm{BA}}$ to translate an image style from domain B back into domain A. An adversarial loss is applied on the output of each generator: discriminator $D_{\mathrm{B}}$ on the output of generator $G_{\mathrm{AB}}$, and discriminator $D_{\mathrm{A}}$ on the output of generator $G_{\mathrm{BA}}$. The adversarial losses are formulated as:
\begin{equation}
 \mathcal{L}_{\mathrm{B_{adv}}}= (D_{\mathrm{B}}(G_{\mathrm{AB}}(I_{\mathrm{A}}))-1)^2
\end{equation}
\begin{equation}
 \mathcal{L}_{\mathrm{A_{adv}}}= (D_{\mathrm{A}}(G_{\mathrm{BA}}(I_{\mathrm{B}}))-1)^2
\end{equation}

The complete adversarial objective to be minimized $\mathcal{L}_{\mathrm{adv}}$ is:
\begin{equation}
 \mathcal{L}_{\mathrm{adv}}= \mathcal{L}_{\mathrm{B_{adv}}} + \mathcal{L}_{\mathrm{A_{adv}}}
\end{equation}

We train the discriminators to minimize the following objective:
\begin{equation}
 \mathcal{L}_{\mathrm{B_{disc}}}= (D_{\mathrm{B}}(I_{\mathrm{B}})-1)^2 + (D_{\mathrm{B}}(G_{\mathrm{AB}}(I_{\mathrm{A}})))^2
\end{equation}
\begin{equation}
 \mathcal{L}_{\mathrm{A_{disc}}}= (D_{\mathrm{A}}(I_{\mathrm{A}})-1)^2 + (D_{\mathrm{A}}(G_{\mathrm{BA}}(I_{\mathrm{B}})))^2
\end{equation}

The complete discriminator objective to be minimized $\mathcal{L}_{\mathrm{disc}}$ is:
\begin{equation}
 \mathcal{L}_{\mathrm{disc}}= \mathcal{L}_{\mathrm{B_{disc}}} + \mathcal{L}_{\mathrm{A_{disc}}}
\end{equation}

\begin{figure}
\centering
\noindent\includegraphics[width=1.0\columnwidth]{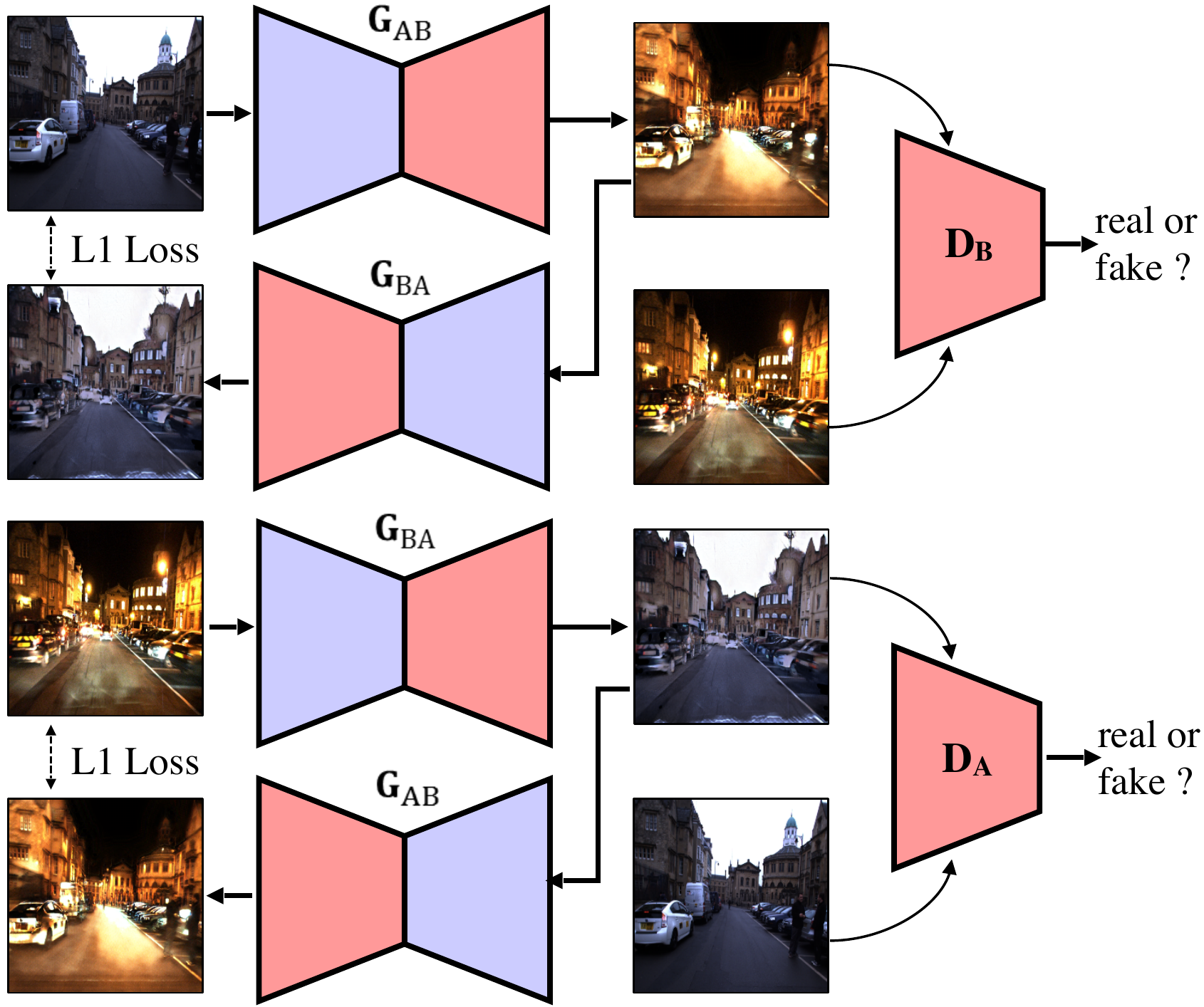} 
\caption{The overall training architecture of the cycle-consistency GAN used to generate synthetic training data. We follow the training regimen described in \cite{CycleGAN2017} for each of the $N$ pairings between the \textbf{reference} condition and a \textbf{target} condition. }
\label{fig:cyclearch}
\end{figure}

A cycle-consistency loss \cite{CycleGAN2017} is applied between the reconstructed and input images: 
\begin{equation}
 \mathcal{L}_{\mathrm{rec}}={\lVert I_{\mathrm{input}} - I_{\mathrm{reconstructed}} \rVert}_{1}
\end{equation}
The final generator objective $\mathcal{L}_{\mathrm{gen}}$ is:
\begin{equation}
 \mathcal{L}_{\mathrm{gen}}=\lambda_{\mathrm{rec}}*\mathcal{L}_{\mathrm{rec}} + \lambda_{\mathrm{adv}} * \mathcal{L}_{\mathrm{adv}}
\end{equation}

with each $\lambda$ term representing a hyperparameter that weighs the importance of each individual objective. We want to find the optimal generators $G_{\mathrm{AB}}$, $G_{\mathrm{BA}}$ that minimize the complete objective:
\begin{equation}
 G_{\mathrm{AB}}, G_{\mathrm{BA}} = \underset{G_{\mathrm{AB}},G_{\mathrm{BA}},D_{\mathrm{B}},D_{\mathrm{A}}}{\arg\min} \mathcal{L}_{\mathrm{gen}} + \mathcal{L}_{\mathrm{disc}}
\end{equation}
We follow this methodology for $N$ difficult conditions (domain B), always paired with the \textbf{reference} condition (domain A), yielding $2N$ generators. However, once the generators have converged, we only use the generators that apply the style of domains B to images from domain A - $G_{\mathrm{AB}}$ - to generate $N$ versions of the \textbf{reference} sequence, each bearing the appearance of a sequence from domain B. Please note that for brevity we omit the condition-specific subscripts from the equations above. An overview of the CycleGAN architecture is shown if Figure \ref{fig:cyclearch}.


\subsection{Input Adapters}\label{subsec:cleaner}
The second step in our approach is to use the data generated in the previous step to train a bank of adapters that preprocess the input images such that they follow a distribution similar to that of the training sets used to train the bank of tasks.
We formulate our input adapters as convolutional encoder-decoders with 3 down-convolutions, a bottleneck with $N_{res}$ ResNet \cite{resnet} blocks and 3 transpose-convolutions. The input to our adaptors is a 3-channel RGB image, while the output is a 3-channel image compatible with the inputs of many well-known models (semantic segmentation, object detection, depth estimation etc). This configuration provides a light-weight solution that is easy to train using labelled data, with reduced storage requirements and a small run-time memory footprint.

For each input adapter $F_{\mathrm{k}}$ (specific to a $k^{th}$ particular appearance), and each task $T_{m}$, we formulate the following loss:
\begin{equation}
 \mathcal{L}_{T_{m}} = T_{m}(F_{\mathrm{k}}(G_{AB_{k}}(I_A))) - T_{m}(I_A)
\end{equation}
where $G_{AB_{k}}$ is the CycleGAN generator that transforms images from the \textbf{reference} condition to the $k^{th}$ condition, and $I_{A}$ is an input \textbf{reference} image. For each input adapter $F_{\mathrm{k}}$ specific to condition $k$ out of $N$ conditions, and $M$ tasks, the final objective to be optimized becomes:
\vspace{-2mm}
\begin{equation}
 F_{\mathrm{k}} = \underset{F_{\mathrm{k}}}{\arg\min} 
 \sum_{m=1}^{M} \alpha_{m} * \mathcal{L}_{T_{m}}
\end{equation}
where $\alpha_{m}$ is a non-negative weight modulating the importance of each task $T_{m}$.

One key takeaway here is that $F_{\mathrm{k}}$ is essentially different from $G_{BA_{k}}$ (the domain-specific generator that maps an image back to the appearance of the \textbf{reference} sequence) - we are not directly concerned with obtaining images that possess the appearance of the reference sequence, instead we want to obtain a processed image that maximizes the performance on the set of tasks $T_{m}$. This can be observed in Figure \ref{fig:qualitative}, second image from left.
\begin{figure}
\centering
\noindent\includegraphics[width=1.0\columnwidth]{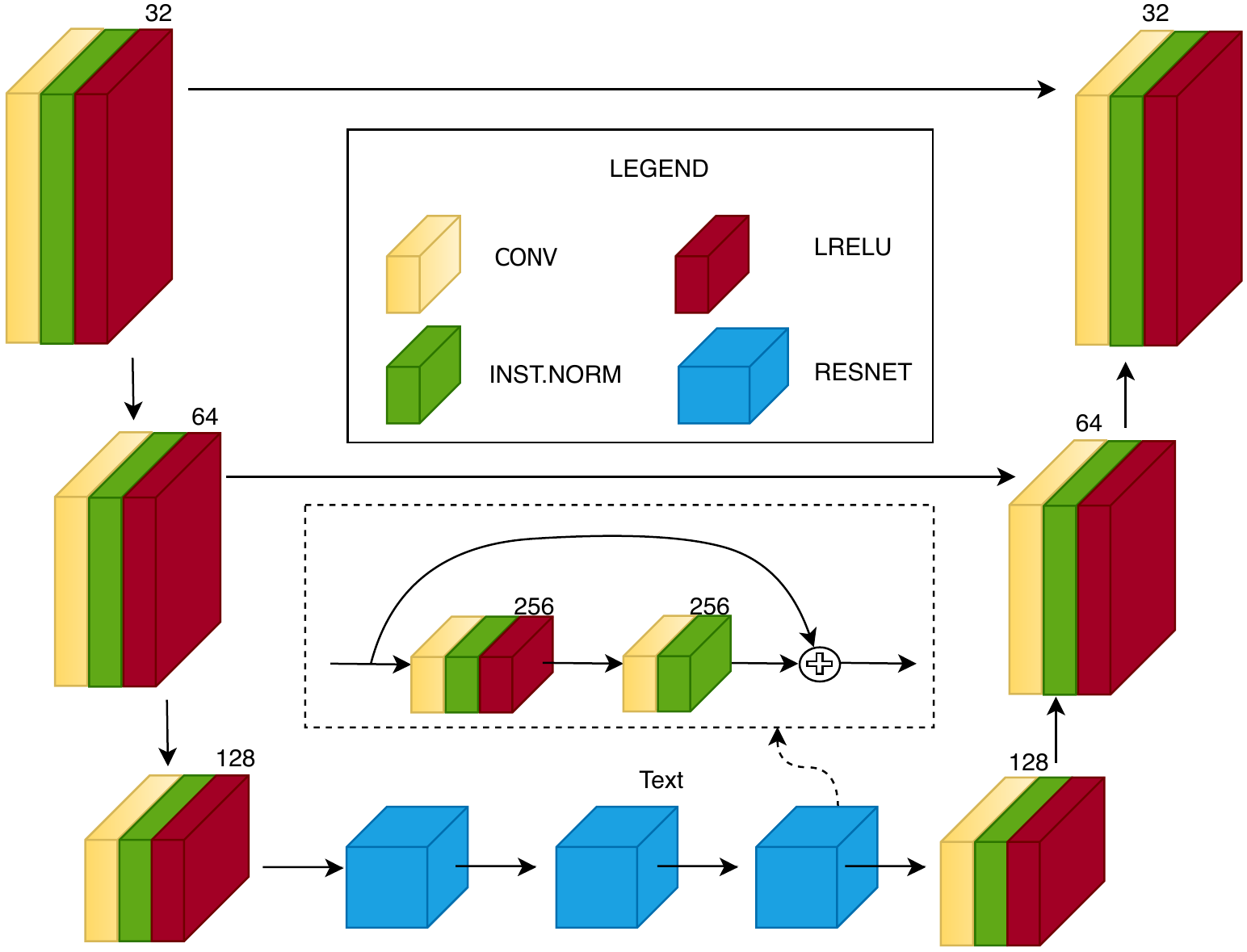} 
\caption{The overall architecture of our input adapters. Under the assumption that a change in condition should not change the overall structure of the scene, we make use of skip-connections \cite{unet} and a ResNet \cite{resnet} bottleneck to facilitate the direct transfer of features from the input side of the network to the output side.}
\label{fig:adapterarch}
\end{figure}

\subsection{Domain Classifier}\label{subsectionclassifier}
We employ a domain classifier $D$ to select the most suitable input adapter $F_{k}$ that enables optimal performance on input images with the $k^{th}$ condition. The classifier follows a largely traditional architecture comprised of 4 convolutional layers and 3 fully connected layers followed by a softmax layer, outputting an $N$-length vector. Given an input image $I_{A}$ and a domain label $t$ as an $N$-length one-hot encoding, we wish to find the parameters of the classifier $D$ that minimizes the cross-entropy between the output of the classifier and the target label $t$:
\vspace{-3mm}
\begin{equation}
 D = \underset{D}{\arg\min} - \sum_{k=1}^{N} t_{k} \cdot \log(D(I_{A})_{k})
\end{equation}
with $k$ used to denote the element in each one-hot encoding.
After training the classifier with $N$ conditions, we additionally use the output of the penultimate fully-connected layer as a length-$128$ condition descriptor, which allows us to discriminatively identify domains that are outside of the original $N$ domains used during training. To do this, we average the descriptors over a sequence of input images, and compare to other stored descriptors in the Euclidean space. A detailed explanation of how this is used is presented in Subsection \ref{subseconlinelearn}.

\begin{figure}
\centering
\noindent\includegraphics[width=1.0\columnwidth]{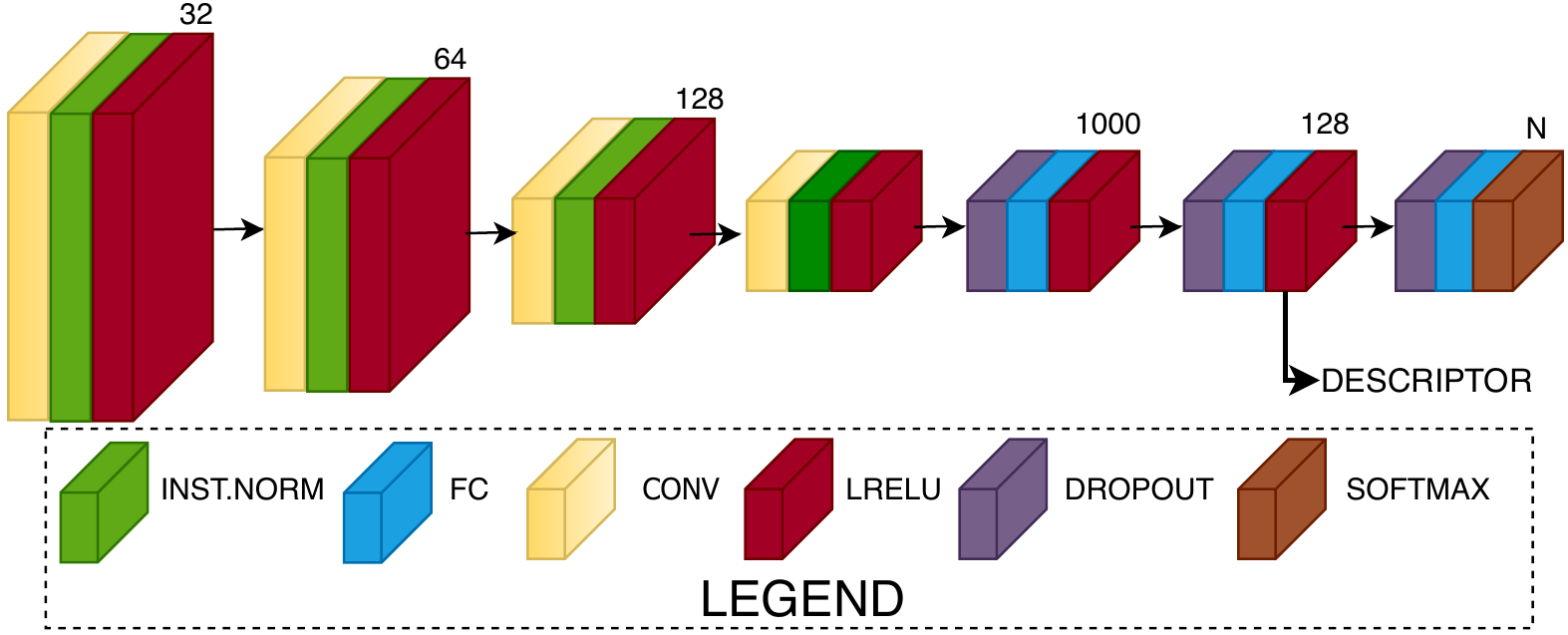} 
\caption{Our classifier follows a traditional architecture, being composed of a series of down-convolutional layers followed by three fully connected layers. For on-line identification of new, unseen domains, we interpret the output of the penultimate fully-connected layer as a discriminative descriptor. Doing so allows us to identify an arbitrary number of domains, beyond the original $N$ domains, without re-training the classifier.}
\label{fig:classifierarch}
\vspace{-5mm}
\end{figure}

\subsection{Parameter Memory}

After training each of the $N$ condition-specific input adapters $F_{k}$ parameters (weights) are stored in a memory (database) $S$. Additionally, the parameters of input adapters fine-tuned following the approach described in Subsection \ref{subseconlinelearn} are also stored in the same memory. The classifier described in Subsection \ref{subsectionclassifier} is used to select a set of optimal parameters to be used in the input adapter $F_{k}$. The memory can be queried in two ways: either by using an index $k$ between $1$ and $N$ (the number of initial conditions) or by specifying a length-$128$ query descriptor and retrieving the set of parameters associated with the descriptor that is closest in the Euclidean space. To enable online learning of unseen domains, we additionally save the parameters of the cycle-consistency GAN generators using the same addressing scheme.

\begin{figure}[htbp]
\vspace{-2mm}
\centering
\noindent\includegraphics[width=1.0\columnwidth]{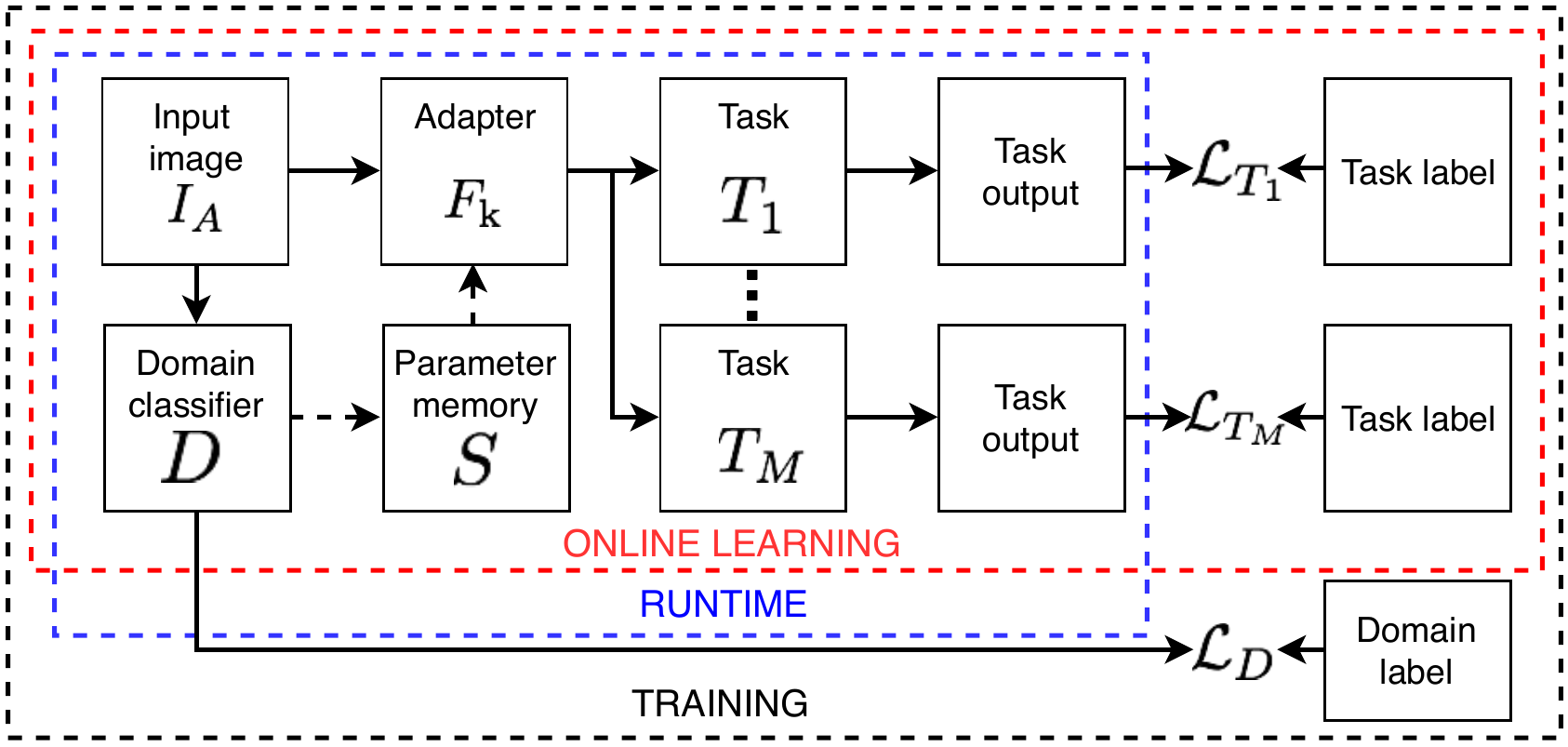} 
\caption{An overview of our train- and test-time pipeline architecture. Given in input image $I_{A}$, the output of the classifier $D$ is used to select a set of parameters for the input adapter $F_k$. The input adapter is then used to transform the input image into a representation that is better suited for the bank of $M$ tasks $T_m$. During training, the performance of the tasks on the transformed input image is used as a corrective signal through the set of losses $\mathcal{L}_{T_{m}}$, along with a domain classification loss $\mathcal{L}_{D}$ used on the output of the classifier. At runtime only the components contained within the blue dotted rectangle are used. During online learning, the components contained within the red dotted rectangle are used. Solid arrow lines represent differentiable paths, while dotted arrow lines represent non-differentiable paths. }
\label{fig:classifierarch}
\vspace{-5mm}
\end{figure}

\subsection{Online Learning}\label{subseconlinelearn}
The pipeline described in the previous subsections can be extended to incremental, unsupervised, online learning of new, unseen domains without requiring any significant modifications to the existing system. We summarize and outline below the processed used:
\begin{itemize}
\item Given a continuous sequence of incoming images, we store the current frame and $T-1$ past frames in a buffer of length $T$ that gets updated using a First-In-First-Out scheme
\item For each frame in the buffer, we compute a length-128 condition descriptor using the penultimate layer of the classifier and average all the descriptors, yielding one single length-$128$ average descriptor
\item  If this average descriptor condition differs(in Euclidean space) by more than a threshold from the descriptors of any conditions previously trained on (i.e. the parameter memory $S$ is unable to reliably identify the condition), the following training pipeline is triggered:
\item  We select the cycle-consistency GAN models closest to the current condition(using the condition descriptor), clone and fine-tune them for the current condition using the sequence stored in the buffer
\item  We use the newly trained generators from above to apply the new style to the \textbf{reference} condition to create a new training sequence
\item  We select the input adapter that is closest to the new condition(again using the descriptor), clone it and train it using the newly created training sequence from above
\item  We begin using this new adapter in the pipeline until the input condition changes significantly again
\end{itemize}

In the following section we describe our experimental setup.
\vspace{-3mm}
\section{Experimental Setup}\label{sec:experimental-setup}

\subsection{Creating multiple conditions}\label{sec:multicond}

From the RobotCar Dataset \cite{RobotCarDatasetIJRR}, we choose $N=7$ initial conditions: Snow, Dusk, Night, Night(rain), Night(low exp.), Shadows, Sun(glare) and a \textbf{reference} condition with a daytime, overcast condition. Additionally, we choose Sun (with ultrahigh exposure) as a condition not seen during initial training, to be used for online training.

\subsection{Training}
For training the cycle-consistency GAN models, we closely follow the approach from \cite{CycleGAN2017}, and the reader is encouraged to consult the publication for further details. We train each of the initial $N$ condition pairs for 100 epochs, and each online fine-tuning stage for 5 epochs. 
For training the input adapters, we use the Adam optimizer \cite{kingma2014adam} with a base learning rate of 0.0005, $\beta_{1}=0.9$ and $\beta_{2}=0.999$. We have found that training an initial adapter on the \textbf{reference} condition (creating an identity function) and then using the parameters of this adapter as a 'seed' for training the initial $N$ adapters greatly stabilizes and speeds up training. This provides the hint that most parametrisations for different condition adapters lie relatively close to each other on the parameter manifold, partly explaining the relative efficiency of our approach to incremental domain adaptation. We train the input adapters for 20 (offline) or 5 (online) epochs or until performance on the validation split stops increasing, whichever comes first.

\subsection{The tasks}
For our particular experiment, we chose $M=2$ tasks: semantic segmentation and topological localisation, since they represent two critical components for robotics. For the semantic segmentation task, we chose to use DeepLab V3+, as it produces state of the art results on a number of standard benchmarks \cite{deeplabv3plus2018}. We use a model \fnurl{checkpoint}{https://github.com/tensorflow/models/blob/master/research/deeplab/g3doc/model_zoo.md}  trained on the Cityscapes dataset \cite{cordts2016cityscapes} that achieves 0.83\% mIOU on the Cityscapes test split. For topological localisation, we chose the de-facto standard approach of computing place descriptors, NetVLAD \cite{netvlad}, and used L2 matching. We use a model  \fnurl{checkpoint}{https://github.com/uzh-rpg/netvlad_tf_open} trained on the Pitts30K dataset \cite{Pitts30k}.

Both of these architectures may be freely swapped with others as long as they are end-to-end differentiable. We set $\lambda=1$ for semantic segmentation and $\lambda=10$ for topological localisation, as we have noticed that increasing the importance of the localisation task improves overall performance without affecting the semantic segmentation task.
\vspace{-2mm}
\subsection{Performance}
The input adapter performs inference at approximately $20$\,Hz for RGB inputs with a size of $640\times480$, on an Nvidia Titan V GPU. The chosen tasks have independent runtime performances of $3$\,Hz and $20$\,Hz for semantic segmentation and topological localisation, respectively.
 
We benchmark on-line learning by introducing an unseen condition (Sun with ultrahigh camera exposure). When starting from a system trained on $N=7$ initial conditions described above, with the closest condition being Sun with glare, on-line training for the new domain takes approximately 30 minutes to first fine-tune the cycle-consistency GAN generators, followed by approximately 10 minutes to train the input adapter. This gives a complete cycle of 40 minutes, meaning that we can fine-tune for new domains approximately 36 times per day. This time should decrease with the addition of more conditions, as new domains could then benefit from 'closer' seeds when performing online training.
\vspace{-4mm}
\section{Results}\label{sec:results}

For semantic segmentation, we create a testing split from the RobotCar Dataset \cite{RobotCarDatasetIJRR} \textbf{reference} sequence and generate testing sequences with different conditions by applying style-transfer using the cycle-consistency GAN generators obtained during the training stage of our pipeline. This process yields sequences with $8$ different conditions (the $7$ initial conditions and one additional condition for testing online learning) and a common approximated ground truth. We again wish to remind the reader that testing is done on sequences derived from the \textbf{reference} sequence through style transfer (along with the approximated ground truth) due to a lack of semantic annotation for the RobotCar Dataset. 
To show the increase in segmentation performance on a dataset with hand-labelled groundtruth, we further test our system on the validation split of the BDD100K segmentation dataset \cite{BDD100K}. As we train exclusively on data from the RobotCar Dataset, BDD is a domain that has \emph{never} been seen during training of any components of our pipeline, and better reflects the usefulness of the proposed pipeline. As the BDD validation sequence does not have enough instances of each condition to also demonstrate online-learning, we freeze our system trained on $8$ conditions ($N=7$ initial conditions and $1$ online-learned condition) and test it on BDD.
For topological localisation, we test on \textbf{real} sequences (not produced using style-transfer) from the RobotCarDataset using the provided INS-RTK GPS ground truth, using a tolerance of $5$ meters and reporting Precision-Recall and Area Under Curve (AUC).
\vspace{-2mm}
\subsection{Quantitative results}

\begin{figure*}[htbp]
\noindent\includegraphics[width=1.0\textwidth]{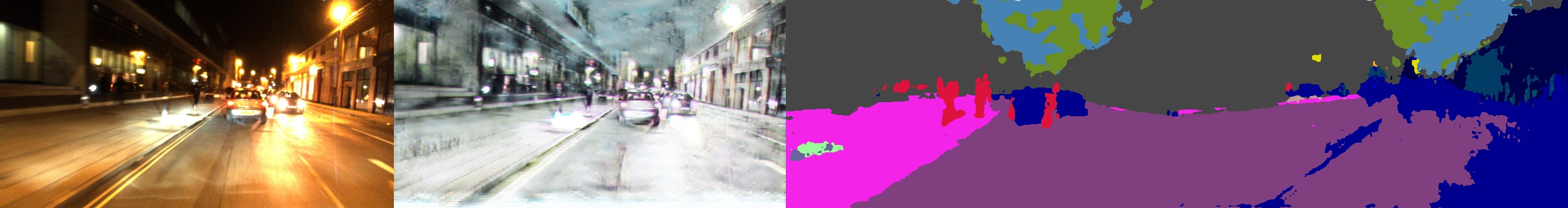}
\vspace{-7mm}
\caption{Improvement of semantic segmentation on a \textbf{real} RobotCar night-time input. The first image is the input image, the second image is the output of the selected adapter, the third image is the result of running the segmentation model on the adapted image, while the last image is the result of running the segmentation model on the original, raw input image.}
\label{fig:qualitative}
\vspace{-3mm}
\end{figure*}

Semantic segmentation performance is significantly improved for RobotCar sequences, as can be seen in Table \ref{rc-mious}. We compare our method (Indiv. adapters) of selecting input adapters with 3 other scenarios: a Baseline where the segmentation model is applied directly on the input image, one where we fine-tune the DeepLab segmentation model on all existing conditions (Deeplab-all) and one where we fine-tune individual DeepLab segmentation models for each condition, and use the classifier output to select the right model. The results show that our method consistently and significantly surpasses all other methods, with an average improvement of over 20 percentage points over the Baseline method. Additionally, night-time conditions show impressive gains in performance, with over 47 percentage points gained over Baseline for the \textbf{Night(rain)} condition.
Table \ref{fig:BDDsegm} presents results for semantic segmentation on the BDD dataset, which has never been seen during training. We test against the same 3 methods described above and again observe significant improvements, with over 5 percentage points gained in Mean Intersection Over Union compared to the Baseline method.

Similarly, topological localization shows a significant overall improvement, with an average of 10 percentage points of overall improvement in Area Under Curve (AUC), and very large improvements for Sun(glare) and Shadows. Night-time traversals are one exception where the improvements are still positive but smaller, as the task of detecting discriminative features is arguably harder than performing segmentation. Table \ref{rc-topo-aucs} and Figures \ref{fig:PRWITH},\ref{fig:PRWITHOUT} present the results in more detail.

\begin{figure}[htbp]
\vspace{-5mm}
\noindent\includegraphics[width=1.0\columnwidth]{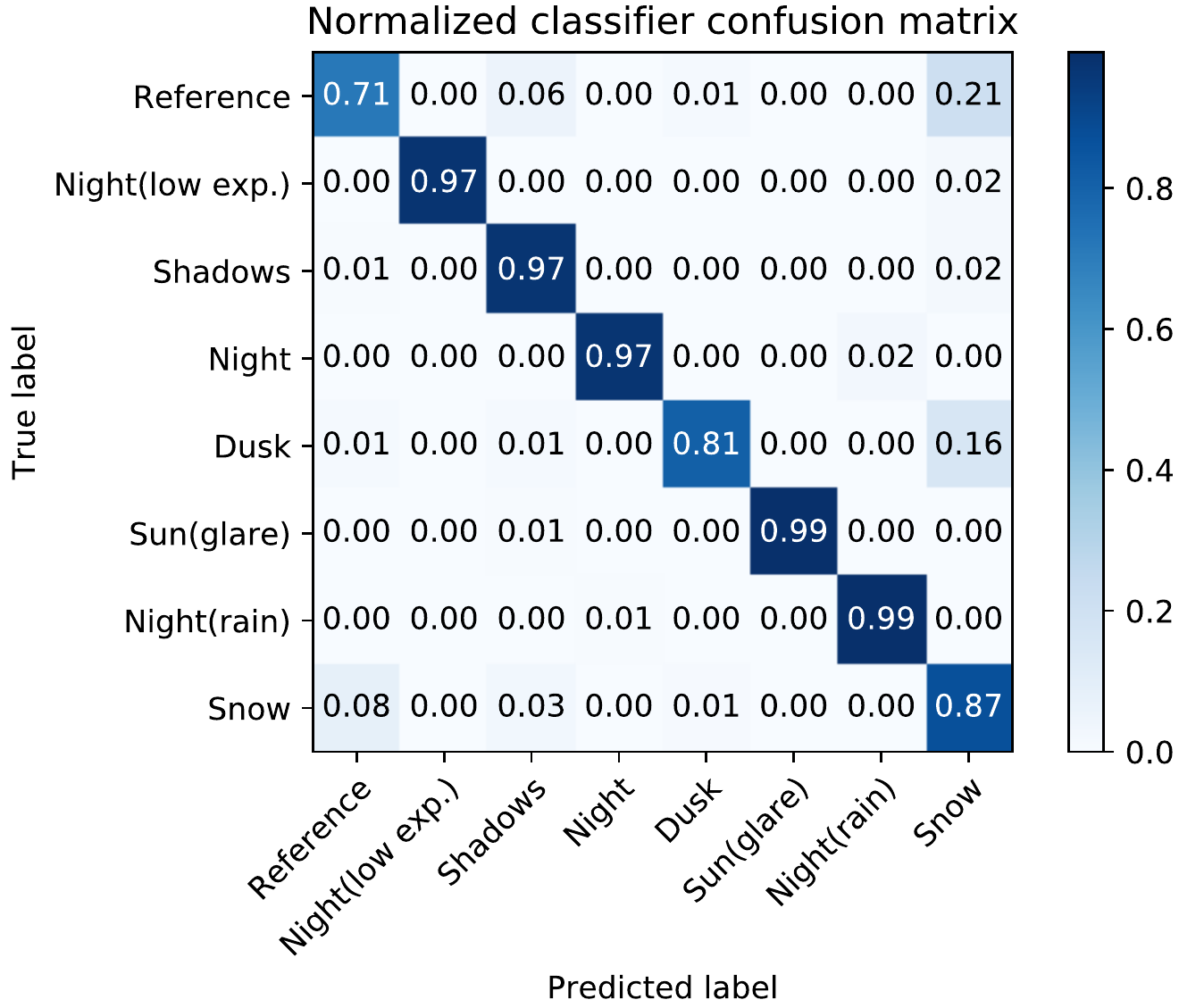} 
\vspace{-7mm}
\caption{RobotCar Condition Classifier confusion matrix. Our condition classifier achieves a 91\% overall accuracy rate.}
\label{fig:confmatrix}
\vspace{-4mm}
\end{figure}

The condition classifier has an overall accuracy of 91\%. The classifier confusion matrix is presented in Figure \ref{fig:confmatrix}. The confusion of the reference, dusk and snow conditions does not lead, empirically, to a large drop in upstream task performance as there is a large degree of similarity between them.

\subsection{Qualitative results}
Additionally, we inspect segmentation results on \textbf{real} sequences (not produced using style-transfer) from the RobotCar Dataset. While they possess the same range of conditions as the ones the system was trained on, a ground truth is not available. We observe improvements in segmentation across the board, with the most important classes (vehicles, pedestrians, bicyclists etc) becoming distinguishable in even the most difficult conditions. An example for night-time is given in Figure \ref{fig:qualitative}.

\subsection{Online learning results}
\begin{figure}[htbp]
\vspace{-3mm}
\noindent\includegraphics[width=1.0\columnwidth]{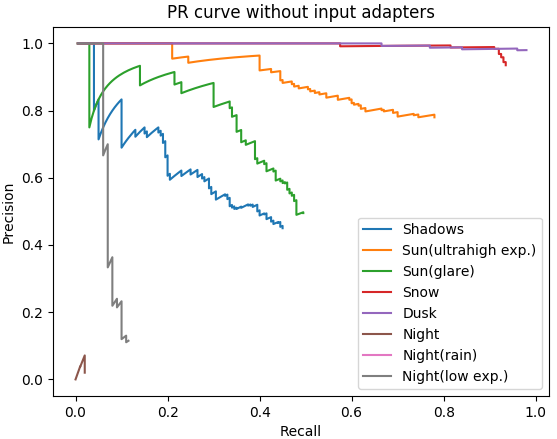}
\vspace{-5mm}
\caption{RobotCar topological localisation Precision-Recall without input adapters. The AUC values can be found in Table \ref{rc-topo-aucs}.}
\label{fig:PRWITHOUT}
\end{figure}

\begin{figure}[htbp]
\vspace{-5mm}
\noindent\includegraphics[width=1.0\columnwidth]{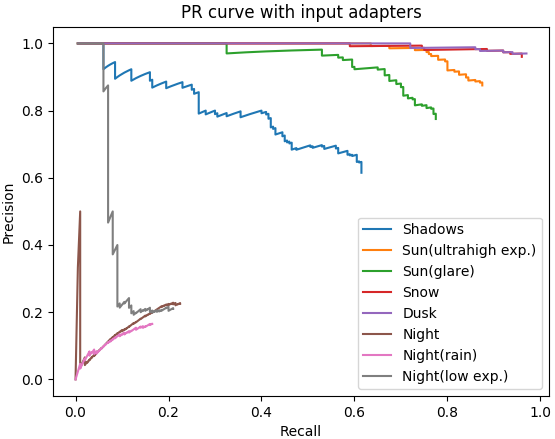} 
\vspace{-7mm}
\caption{RobotCar topological localisation Precision-Recall with input adapters. The AUC values can be found in Table \ref{rc-topo-aucs}.}
\label{fig:PRWITH}
\end{figure}

To test our online learning capabilities, we run our system on a condition never before seen during training, \textbf{Sun(with ultrahigh camera exposure)}. The descriptor extracted from the penultimate layer of the classifier cannot be accurately matched to any stored descriptors, so the online training process is triggered. In the descriptor feature space, the closest condition stored is \textbf{Sun(glare)}, which is used as a seed for training the cycle-consistency GAN generators and a new input adapter. Results for this new condition are presented in Figures \ref{fig:PRWITHOUT}, \ref{fig:PRWITH} and in the \emph{gray} shaded columns in Tables \ref{rc-topo-aucs} and \ref{rc-mious}. As with the initial condition, we observe a large and significant increase in performance for both topological localisation (over 15 percentage points) and semantic segmentation (11 percentage points).

\begin{table}[!htbp]\centering
\vspace{2mm}
\caption{BDD segmentation Mean Intersection Over Union (mIOU)}
\vspace{-3mm}
\begin{tabular}{|c|c|c|c|}
\hline
No adapter & Deeplab-all & Indiv. Deeplabs  & Indiv. adapters(\textbf{ours}) \\ \hline
0.4070  & 0.4100   & 0.4319                   & \textbf{0.4503}   \\ \hline
\end{tabular}
\label{fig:BDDsegm}
\vspace{-5mm}
\end{table}

\begin{table*}[!htbp]\centering
\caption{RobotCar topological localisation Area Under Curve (AUC)}
\vspace{-3mm}
\begin{tabular}{|c|c|c|c|c|c|c|c|>{\columncolor[gray]{0.85}}c|c|}
\hline
Method          & Shadows     & Sun(glare)  & Snow        & Dusk        & Night       & Night(rain) & Night(low exp.) & Sun(ultrahigh exp.) & Mean \\ \hline
No adapters & 0.2928  & 0.3949 & 0.9263 & \textbf{0.9710} & 0.0009 & 0.0000           & 0.0719  & 0.7043 & 0.4202   \\ \hline
With adapters(\textbf{ours})    & \textbf{0.4965}  & \textbf{0.7404} & \textbf{0.9494} & 0.9605 & \textbf{0.0374} & \textbf{0.0191} & \textbf{0.0981} & \textbf{0.8593}  & \textbf{0.5200}     \\ \hline
\end{tabular}
\label{rc-topo-aucs}
\vspace{-2mm}
\end{table*}

\vspace{-3mm}
\begin{table*}[!htbp]\centering
\caption{RobotCar segmentation Mean Intersection Over Union (mIOU)}
\vspace{-3mm}
\begin{tabular}{|c|c|c|c|c|c|c|c|>{\columncolor[gray]{0.85}}c|c|}
\hline
Method & Shadows & Sun(glare) & Snow & Dusk & Night & Night(rain) & Night(low exp.) & \textit{Sun(ultrahigh exp.)} & Mean \\ \hline
Baseline & 0.6014 & 0.4316 & 0.5677 & 0.6156 & 0.1404 & 0.0859 & 0.1850 & \textit{0.4423} & 0.3837 \\ \hline
Deeplab-all & 0.5375 & 0.4712 & 0.5055 & 0.5372 & 0.3465 & 0.3572 & 0.3593 & \textit{0.4821} & 0.4495 \\ \hline
Indiv. deeplabs & 0.5594 & 0.4948 & 0.5303 & 0.5656 & 0.3948 & 0.4138 & 0.4184 & \textit{0.5120} & 0.4861 \\ \hline
Indiv. adapters(\textbf{ours}) & \textbf{0.6292} & \textbf{0.6419} & \textbf{0.6525} & \textbf{0.6327} & \textbf{0.5301} & \textbf{0.5627} & \textbf{0.5136} & \textit{\textbf{0.5500}} & \textbf{0.5891} \\ \hline
\end{tabular}
\label{rc-mious}
\vspace{-5mm}
\end{table*}

\vspace{-1mm}
\section{Conclusions}\label{sec:conclusions}

To prevent performance of computer vision tasks from degrading quickly and often catastrophically when input conditions change, we have presented a domain adaptation system that uses light-weight input adapters to pre-processes input images, irrespective of their appearance, in a way that makes them compatible with off-the-shelf computer vision tasks that are trained only on inputs with ideal conditions. No fine-tuning is performed on the off-the-shelf models, and the system is capable of incrementally training new input adapters in a self-supervised fashion, using the computer vision tasks as supervisors, when the input domain differs significantly from previously seen domains. We report large improvements in semantic segmentation and topological localization performance on two popular datasets, RobotCar and BDD. This work is presented as a framework, and each end-to-end differentiable component may be replaced with a better-performing counterpart, or with one that is better-suited for the task at hand, if available. Additionally, the training process may be extended to work with an arbitrary number of supervisory signals. Finally, our on-line training regimen benefits from convergence times that decrease as a function of the number of domains trained on.

\vspace{-2mm}
\section{Acknowledgements}\label{sec:acknowledgements}
\vspace{-1mm}
This work was supported by a Oxford-Google DeepMind Graduate Scholarship and EPSRC/UK Research and Innovation Programme Grant EP/M019918/1.

\vspace{-2mm}
\bibliographystyle{IEEEtran}
{\footnotesize
\bibliography{biblio}}

\end{document}